\documentclass[10pt,twocolumn,letterpaper]{article}

\usepackage{btas}
\usepackage{times}
\usepackage{epsfig}
\usepackage{graphicx}
\usepackage{amsmath}
\usepackage{amssymb}
\usepackage{multirow}
\usepackage[norule,symbol,perpage]{footmisc}

\usepackage[pagebackref=true,breaklinks=true,letterpaper=true,bookmarks=false]{hyperref}

\btasfinalcopy 

\ifbtasfinal\fi

\makeatletter
\def\ps@IEEEtitlepagestyle{\def\@oddfoot{\mycopyrightnotice}
\def\@evenfoot{}} 
\def\mycopyrightnotice{{\hfill \footnotesize 978-1-7281-1522-1/19/\$31.00 \copyright 2019 IEEE\hfill}}
\makeatother

\begin{document}

\title{Subclass Contrastive Loss for Injured Face Recognition}

\author{Puspita Majumdar, Saheb Chhabra, Richa Singh, Mayank Vatsa\\
IIIT-Delhi, India\\
{\tt \{pushpitam, sahebc, rsingh, mayank\}@iiitd.ac.in}
}

\maketitle
\thispagestyle{empty}

\begin{abstract}
Deaths and injuries are common in road accidents, violence, and natural disaster. In such cases, one of the main tasks of responders is to retrieve the identity of the victims to reunite families and ensure proper identification of deceased/ injured individuals. Apart from this, identification of unidentified dead bodies due to violence and accidents is crucial for the police investigation. In the absence of identification cards, current practices for this task include DNA profiling and dental profiling. Face is one of the most commonly used and widely accepted biometric modalities for recognition. However, face recognition is challenging in the presence of facial injuries such as swelling, bruises, blood clots, laceration, and avulsion which affect the features used in recognition. In this paper, for the first time, we address the problem of injured face recognition and propose a novel \textbf{Subclass Contrastive Loss (SCL)} for this task. A novel database, termed as \textbf{Injured Face (IF)} database, is also created to instigate research in this direction. Experimental analysis shows that the proposed loss function surpasses existing algorithm for injured face recognition.

\end{abstract}

\let\thefootnote\relax\footnotetext{\mycopyrightnotice}

\begin{figure}[t]
\centering
\includegraphics[scale = 0.45]{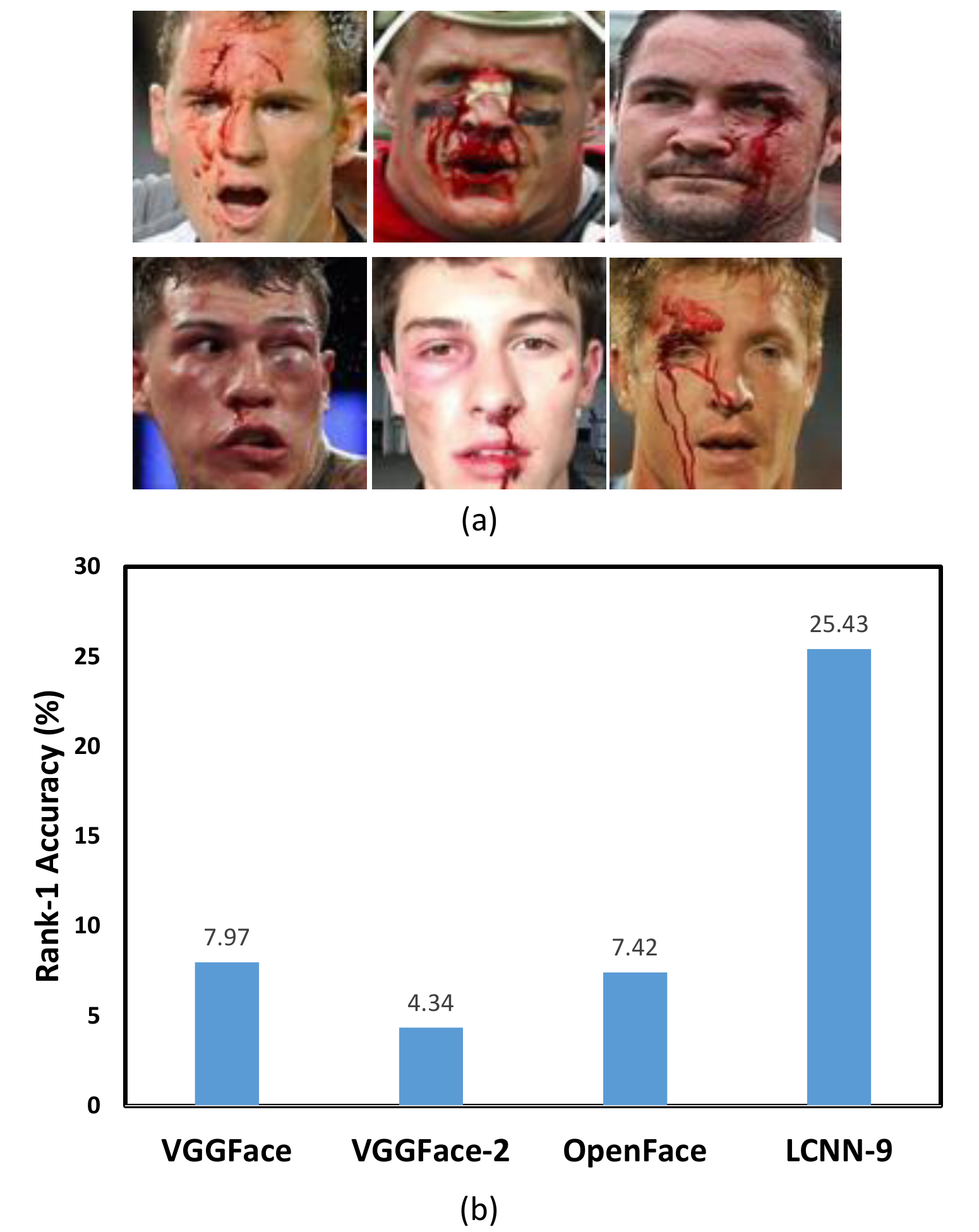}
\caption{Illustrating the impact of injured faces on existing face recognition models. (a) Sample images showing the change in appearance due to injuries (images are taken from Internet). (b) Performance of existing face recognition models for identifying injured faces at rank-1 (experimental details are available in Section 5.1).}
\label{fig:Visual_Abstract}
\end{figure}

\section{Introduction}
Injuries on different body parts including face, head, limbs, and neck are common in road accidents, violence, and natural calamities. Among these, face is one of the most affected regions of the human body. As per the report of World Health Organization (WHO) \cite{WinNT1}, every year, 1.25 million people are killed and 50 million are injured in road accidents worldwide. According to Jordan and Calhoun \cite{jordan2006management}, 50\% to 70\% of people surviving traffic accidents have facial injuries. In the majority of such cases, facial regions are partially or completely affected, and identification of victims in such scenarios becomes difficult. The problem becomes more challenging if the victim is in an unconscious state without any identity proofs. Apart from this, many times, natural calamities and violence result in unidentified dead bodies with facial injuries.  Image of unidentified bodies with facial injuries, generally, are not accepted by print or electronic media for publication as they may cause anxiety and discomfort to the viewers \cite{bodkha2012role}. In such situations, retrieving the identity of the deceased person becomes an arduous task. 

\begin{figure*}[t]
\centering
\includegraphics[scale = 0.378]{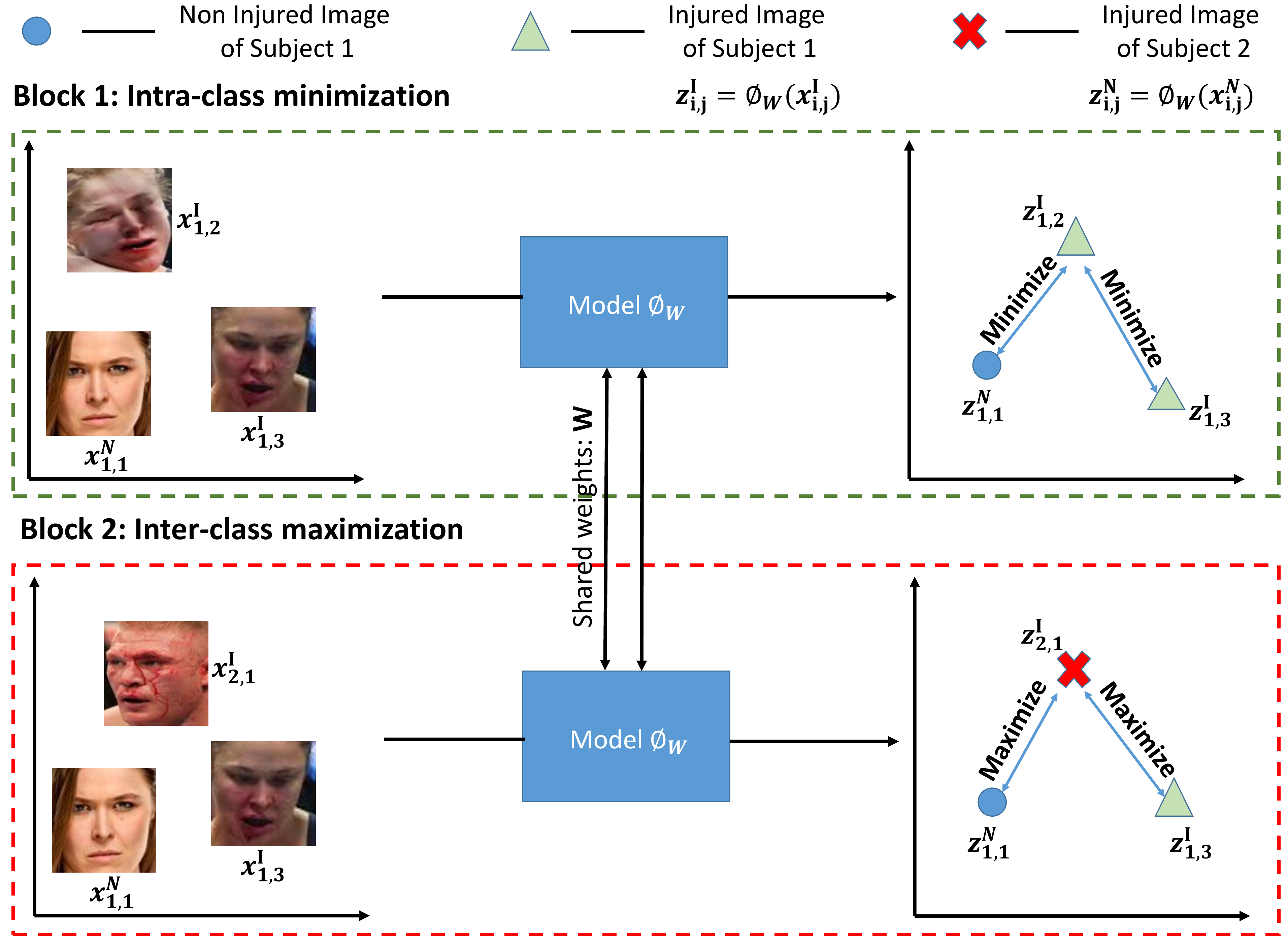}
\caption{Illustrating the steps involved in the proposed approach. During training, when the genuine set is given as an input to the shared model (with weights $\phi_W$), intra-class loss is minimized to reduce the intra-class distance. On the other hand, when imposter set is given as input, inter-class loss is maximized to increase the inter-class separability.}
\label{fig:Block_Diagram}
\end{figure*}

Forensic experts and odontologists use a variety of techniques to ascertain the identity of injured victims and unidentified dead bodies such as DNA profiling and dental profiling \cite{clayton1995identification, quinet2016unclaimed, trokielewicz2016post}. However, these techniques are generally time-consuming, costly, and challenging to scale-up.  In contrast, it is easy to obtain the face images from the victim, which can be matched with the reference images in large (national level) databases. With this motivation, we propose to use automated face recognition for retrieving the identity of unidentified victims and present a solution for injured face recognition. Figure \ref{fig:Visual_Abstract}(a) shows sample images with injured faces. During injuries, facial features are affected due to swelling, bruises, mutilation, and laceration which in turn changes the facial structure. As shown in Figure \ref{fig:Visual_Abstract}(b), existing face recognition models trained on non-injured/normal faces do not perform well for injured face recognition. For matching injured faces with non-injured faces, a novel loss function, termed as \textit{Subclass Contrastive Loss (SCL)}, is proposed. The proposed loss function minimizes the distance between non-injured and injured subclasses of the same subject while maximizes the distance from other subjects. Additionally, a first of its kind \textit{Injured Face (IF)} database is created to evaluate the performance of the proposed loss function. Experiments on the IF database and comparison with existing algorithms showcase the efficacy of the proposed SCL.



\section{Related Work}
In the literature, facial injuries have been studied for various purposes such as identification of dead bodies, distinguishing injuries of domestic violence from accidents, and for studying the psychological effect of facial trauma. Facial features are heavily affected during injuries. This makes the identification of the victim a difficult task.  Different techniques such as face reconstruction or sketch of the victim drawn by sketch artists are used to recover the identities of the victims. Singh et al. \cite{singh2012maxillofacial} evaluated the incidence of maxillofacial injury and concluded that road traffic accidents are the major cause of facial injury. It is found that facial injury is more frequent in male as compared to female. Black et al. \cite{black2017prevention} studied sports injuries and found that more than 41\% of sports injuries require medical treatment. They found that in high-risk sports, eye protection, mouth guards, helmets, and face guard plays a key role in reducing facial injuries. However, these safety measures are not followed in most of the high-risk sports. Canzi et al. \cite{canzi2019cfi} proposed a measuring scale of the facial trauma that summarizes the surgical duration and care required for the treatment. This score can be used in trauma center decision making.

Iris recognition and dental profiling are some of the techniques for identifying dead bodies. In 2016, Trokielewiez et al. \cite{trokielewicz2016post} performed postmortem iris recognition. They evaluated four independent iris recognition methods and found that the irises are correctly recognizable with more than 90 percent accuracy after a few hours of death. Further, Trokielewiez et al. \cite{trokielewicz2019iris} extended their work and collected a database of 1200 near-infrared and 1787 visible light samples of 37 deceased persons. They have shown that iris recognition performs well upto 5-7 hours of death. Nassar et al. \cite{nassar2008automatic} proposed an automated method for postmortem identification by constructing dental charts. Abaza et al. \cite{abaza2009retrieving} proposed a method for the task of efficient retrieval of dental records from a database to assist the forensic experts in identifying deceased individuals in a rapid manner. 


Recently, Coulibaly et al. \cite{coulibaly2018inter} provided a ten-year survey related to interpersonal violence (IPV). It is observed that IPV related facial injuries occur mostly in young and male adults involved in brawls. In 2018, Majumdar et al. \cite{majumdar2018detecting} proposed a framework to distinguish injuries of domestic violence from others. They fine-tuned VGG-Face \cite{parkhi2015deep} model for classification and have shown that the features extracted from the injured regions are useful in classifying victims of domestic violence from others.

\section{Injured Face Recognition using Subclass Contrastive Loss}
Existing deep face recognition models are trained on non-injured or normal faces. However, the feature distribution of regular faces would be different from injured faces. Therefore, existing models, generally, do not perform well for the task of recognition of injured faces (as shown in Figure \ref{fig:Visual_Abstract}). In order to address the problem of injured face recognition, a loss function, termed as Subclass Contrastive Loss (SCL), is proposed which minimizes the distance between subclasses of the same subject and maximizes the distance from subclasses of others. In the majority of the cases, such as, unidentified bodies with facial injuries, injured faces are required to be matched with existing databases of non-injured images for retrieving the identity. Considering these scenarios, in this research, we are assuming injured faces as probe images and non-injured faces as the enrolled gallery. The problem of recognizing injured face images can be formally defined as \textit{``given a dataset $\mathbf{X}$ of non-injured and injured face images of each subject, train a model $\phi_\mathbf{W}$ such that the embeddings of non-injured and injured images of the same subject are close to each other while the embeddings of different subjects are apart in the embedding space''}. Figure \ref{fig:Block_Diagram} shows the block diagram of the proposed approach. 

Let $\mathbf{X}_i$ represents the subject $i$ of dataset $\mathbf{X}$. The set of non-injured images belong to subclass $\mathbf{X}_i^{N}$ and the set of injured images belong to subclass $\mathbf{X}_i^I$. Mathematically, it is represented as:
\begin{equation}
    \mathbf{X}_i = \{\mathbf{X}_i^N, \mathbf{X}_i^I\}
\end{equation}
Let $\phi_\mathbf{W}$ be a Convolutional Neural Network (CNN) with weights $\mathbf{W}$ which outputs the feature representation of the input. The convolutional network $\phi_\mathbf{W}$ is trained in the verification mode. For this purpose, two sets are created, namely genuine and imposter. The genuine set consists of two pairs of samples of the same subject. The first pair is created by taking a sample from the non-injured subclass and another sample from the injured subclass while the second pair is created by taking two different samples from the injured subclass. On the other hand, the imposter set consists of two pairs of samples of different subjects. The first pair contains samples from the non-injured and injured subclass while the second pair contains samples from the injured subclass. Mathematically, a genuine set and an imposter set are represented as:
\begin{equation}
    \mathbf{G} = [\{\mathbf{x}_{i,p}^N, \mathbf{x}_{i,q}^I\}, \{ \mathbf{x}_{i,q}^I, \mathbf{x}_{i,r}^I \}]
\end{equation}
\begin{equation}
    \mathbf{M} = [\{\mathbf{x}_{i,p}^N, \mathbf{x}_{j,q}^I\}, \{ \mathbf{x}_{j,q}^I, \mathbf{x}_{i,r}^I \}]
\end{equation}
where, $\mathbf{x}_{i,p}^N$ is the $p^{th}$ sample of subclass $\mathbf{X}_i^N$ of subject $i$, $\mathbf{x}_{i,q}^I$ and $\mathbf{x}_{i,r}^I$ are $q^{th}$ and $r^{th}$ samples of subclass $\mathbf{X}_i^I$ of subject $i$, and $\mathbf{x}_{j,q}^I$ is the $q^{th}$ sample of subclass $\mathbf{X}_j^I$ of subject $j$. 

\noindent The aim is to minimize the intra-class distance between the embeddings of two subclasses of the same subject and to maximize the inter-class distance from others. In order to achieve this, the network is trained by minimizing the following Subclass Contrastive Loss (SCL) function:
\begin{equation}
    minimize \sum_{G,M}L_{G,M}
\end{equation}
where, $L_{G,M}$ represents the loss for a genuine and imposter set. It is mathematically represented as:
\begin{equation}
    L_{G,M} = (1-Y)L_G^{intra} + YL_M^{inter}
\end{equation}
where, $Y$ is a binary label with $Y$ = 0 for the genuine set and $Y$ = 1 for the imposter set. $L_G^{intra}$ and $L_M^{inter}$ represent the intra-class loss and inter-class loss, which is a distance function of genuine set and imposter set, respectively. For injured face recognition, there are two confounding factors while minimizing the intra-class distance and maximizing the inter-class distance: (i) variations due to the injured and non-injured images and (ii) variations among the injured images (i.e., different types of facial injuries). In order to address the first confounding factor, the distance between the injured and non-injured images of the same subject is reduced, and it is increased for different subjects. Similarly, the second factor is addressed by reducing the distance between the injured images of the same subject and by increasing the distance from others. For any genuine set, loss $L_G^{intra}$ is computed as:
\begin{multline}
    L_G^{intra} = D(\phi_\mathbf{W}(\mathbf{x}_{i,p}^N), \phi_\mathbf{W}(\mathbf{x}_{i,q}^I)) \\+ 
     D(\phi_\mathbf{W}(\mathbf{x}_{i,q}^I),\phi_\mathbf{W}(\mathbf{x}_{i,r}^I)) 
     \label{Eq:intra}
\end{multline}
For an imposter set, loss $L_M^{inter}$ is computed as: 
\begin{multline}     
     L_M^{inter} = max(0,\alpha_1 - D(\phi_\mathbf{W}(\mathbf{x}_{i,p}^N), \phi_\mathbf{W}(\mathbf{x}_{j,q}^I))\\ + max(0,\alpha_2 - D(\phi_\mathbf{W}(\mathbf{x}_{j,q}^I), \phi_\mathbf{W}(\mathbf{x}_{i,r}^I))
     \label{Eq:inter}
\end{multline}
where, $D(\cdot)$ is any distance metric. $\alpha_1$ and $\alpha_2$ are the margins with value greater than 0. $\alpha_1$ controls the separation among the non-injured and injured images of different subjects while $\alpha_2$ controls the separation among the injured images of different subjects. In this research, squared Euclidean distance is used as $D(\cdot)$ and Equations \ref{Eq:intra} and \ref{Eq:inter} are updated as:
\begin{multline}
    L_G^{intra} = \lVert \phi_\mathbf{W}(\mathbf{x}_{i,p}^N) - \phi_\mathbf{W}(\mathbf{x}_{i,q}^I) \rVert_2^2 + \\ \lVert \phi_\mathbf{W}(\mathbf{x}_{i,q}^I) - \phi_\mathbf{W}(\mathbf{x}_{i,r}^I) \rVert_2^2)
\end{multline}
\vspace{-20pt}
\begin{multline}
    L_M^{inter} = max(0,\alpha_1 - \lVert \phi_\mathbf{W}(\mathbf{x}_{i,p}^N) - \phi_\mathbf{W}(\mathbf{x}_{j,q}^I) \rVert_2^2) \\ + max(0,\alpha_2 - \lVert \phi_\mathbf{W}(\mathbf{x}_{j,q}^I) - \phi_\mathbf{W}(\mathbf{x}_{i,r}^I) \rVert_2^2)
\end{multline}
where, $\lVert . \rVert_2^2$ is the squared Euclidean distance.

\begin{figure}[t]
\centering
\includegraphics[scale = 0.35]{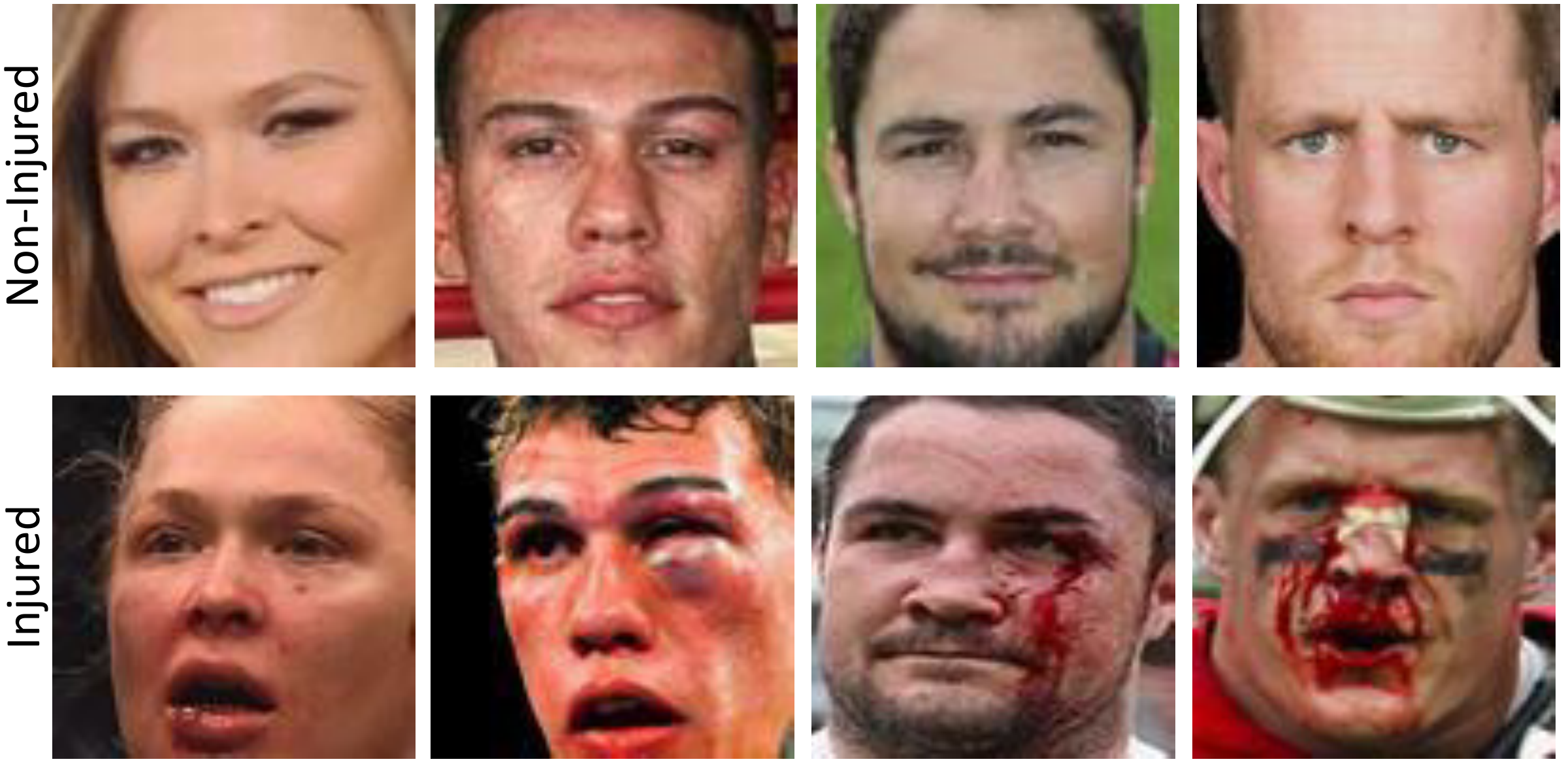}
\caption{Sample images from the Injured Face (IF) database. The first row shows the non-injured images and the second row shows the corresponding injured images.}
\label{fig:DB_Collage}
\end{figure}

\vspace{10pt}

\noindent \textbf{Face Recognition:} For recognition of faces during testing, features corresponding to the non-injured gallery images and injured probe images are extracted using the trained model $\phi_\mathbf{W}$. Extracted features of the probe images are matched with the gallery images using Euclidean distance for identification.

\section{Injured Face Database}
To the best of our knowledge, there is no publicly available database for injured face recognition. Therefore, we have prepared the Injured Face (IF) database by collecting images from online resources. As part of research contributions, the database URLs (i.e., to these images) will be released to the research community. The details of the creation of the IF database and annotation of injured regions are discussed below:

\subsection{Creation of Injured Face Database}
Injured Face (IF) database is created by collecting injured and non-injured/normal face images of different subjects from the Internet. Generally, non-injured/normal face images are used for enrollment. To recognize a subject with facial injuries, face recognition algorithms are required to match the injured face image with the enrolled non-injured images. Therefore, two types of images per subject are collected to create the database: i) faces with real injuries and ii) normal or non-injured face images. The database includes injured face images from sports, accidents, and violence cases. The database contains a total of 898 images corresponding to 100 subjects with 346 injured and 552 non-injured face images. Figure \ref{fig:DB_Collage} shows some sample images of the IF database.  

\subsection{Annotation of Injured Facial Regions}
During facial injuries, different regions of a face are affected. To study the injuries present in different facial regions, injured faces of the IF database are annotated. For this purpose, seven different injured regions of the face are annotated. The regions correspond to forehead, left eye, right eye, left cheek, right cheek, nose, and mouth. Different images of each subject have injuries in one or multiple regions of the face. Table \ref{Annotation} summarizes the number of images with injuries in different facial regions.

\begin{table}[]
\centering
\caption{Number of images with injuries in different facial regions.}
\label{Annotation}
\begin{tabular}{|c|c|c|c|}
\hline
\textbf{Region} & \textbf{Images} & \textbf{Region} & \textbf{Images} \\ \hline
Forehead        & 140             & Right Cheek     & 159             \\ \hline
Left Eye        & 151             & Nose            & 126             \\ \hline
Right Eye       & 124             & Mouth           & 133             \\ \hline
Left Cheek      & 150             & Multiple        & 283             \\ \hline
\end{tabular}
\end{table}

\begin{table}[]
\small
\caption{Details of the experiments performed for Injured Face Recognition (IFR).}
\label{ExptTable}
\centering
\begin{tabular}{|c|c|c|}
\hline
\textbf{Experiment}                                                              & \textbf{Model}                                                                 & \textbf{Database} \\ \hline
\textbf{\begin{tabular}[c]{@{}c@{}}Pre-trained Models\end{tabular}} & \begin{tabular}[c]{@{}c@{}}VGG-Face, OpenFace, \\ VGGFace2, LCNN-9\end{tabular} & IF                \\ \hline
\textbf{Proposed SCL}                                                           & LCNN-9                                                                         & IF                \\ \hline
\textbf{Extended Gallery}                                                        & LCNN-9                                                                         & IF, SCface        \\ \hline
\end{tabular}
\end{table}


\section{Experiments and Results}
To evaluate the performance of existing deep face recognition models and the proposed algorithm with Subclass Contrastive Loss (SCL) on the IF database, multiple experiments are performed. The first experiment is performed for evaluating the performance of four existing pre-trained face recognition models namely, VGG-Face \cite{parkhi2015deep}, OpenFace \cite{amos2016openface}, VGGFace2 \cite{cao2018vggface2}, and LightCNN \cite{wu2018light} on the IF database. The second experiment is performed to evaluate the performance of the proposed algorithm with SCL on the IF database and compared with Contrastive Loss (CL) \cite{hadsell2006dimensionality} and Triplet Loss (TL)  \cite{schroff2015facenet}. Further, an extended gallery experiment is performed to evaluate the robustness of the proposed algorithm by adding images from the SCface database \cite{grgic2011scface} to the gallery of the testing set of the IF database. Table \ref{ExptTable} presents the details of the experiments. Protocol to perform the experiments and implementation details are discussed below:

\noindent \textbf{Protocol:} Experiments are performed by dividing the IF database into training and testing sets with non-overlapping 70\% subjects in the training set and 30\% subjects in the testing set. Five times repeated random sub-sampling is performed for training and testing partitioning. Further, the training and testing sets are divided into gallery and probe, where the gallery contains non-injured/normal images and the probe contains injured images of each subject. Gallery and probe of the training set contain multiple images per subject. In the testing set, the gallery contains a single image per subject while probe contains multiple images per subject. In order to compare the proposed loss function with siamese network trained with Contrastive Loss (CL), pairs are generated using non-injured and injured face images. 

\begin{table}[]
\small
\centering
\caption{Mean classification accuracy (\%) with standard deviation of existing pre-trained models on the IF database.}
\label{BaselineAcc}
\begin{tabular}{|c|c|c|c|}
\hline
                  & \textbf{Rank 1}           & \textbf{Rank 5}           & \textbf{Rank 10}          \\ \hline
\textbf{VGG-Face}  & 7.97 $\pm$ 3.9           & 26.32 $\pm$ 6.0          & 43.70 $\pm$ 6.6          \\ \hline
\textbf{OpenFace} & 7.42 $\pm$ 2.1           & 30.01 $\pm$ 2.6          & 48.93 $\pm$ 5.5          \\ \hline
\textbf{VGGFace2} & 4.34 $\pm$ 0.9           & 19.46 $\pm$ 2.7          & 40.11 $\pm$ 5.2          \\ \hline
\textbf{LCNN-9}     & \textbf{25.43 $\pm$ 6.9} & \textbf{58.67 $\pm$ 6.0} & \textbf{76.65 $\pm$ 7.2} \\ \hline
\end{tabular}
\end{table}

\begin{table}[]
\centering
\small
\caption{Identification accuracy (\%) on the IF database using CL, TL, and SCL.}
\label{ClassAccuracy}
\begin{tabular}{|c|c|c|c|}
\hline
             & \textbf{Rank 1}          & \textbf{Rank 5}          & \textbf{Rank 10}         \\ \hline
\textbf{CL}  & 28.67 $\pm$ 9.2          & 61.72 $\pm$ 6.5          & 80.82 $\pm$ 8.0          \\ \hline
\textbf{TL}  & 30.26 $\pm$ 10.6           & 65.50 $\pm$ 9.9            & 82.15 $\pm$ 5.4            \\ \hline
\textbf{SCL} & \textbf{36.70 $\pm$ 3.7} & \textbf{65.91 $\pm$ 6.0} & \textbf{82.41 $\pm$ 6.5} \\ \hline
\end{tabular}
\end{table}

\noindent \textbf{Implementation Details:} The proposed algorithm is implemented in Pytorch. All the experiments are performed on NVIDIA Tesla P100 server with 96GB RAM and 16GB GPU memory. LightCNN-9 network is used as the base network for the proposed algorithm. Initial two layers of the network are frozen, and the rest of the layers are trained by minimizing the proposed Subclass Contrastive Loss (SCL). Adam optimizer is used with a learning rate of 0.000003. The network is trained for 30 epochs with a batch size of 50. During the experiments, $\alpha_1$ and $\alpha_2$ are set to 2 and 3.1 respectively. LightCNN-9 network with same experimental settings and a margin of 2 is used to train the siamese network with Contrastive Loss (CL) for comparison. LightCNN-9 is also used as the base network for triplet training and the margin is set to 0.4 during the experiments.


\begin{figure}[t]
\centering
\includegraphics[scale = 0.55]{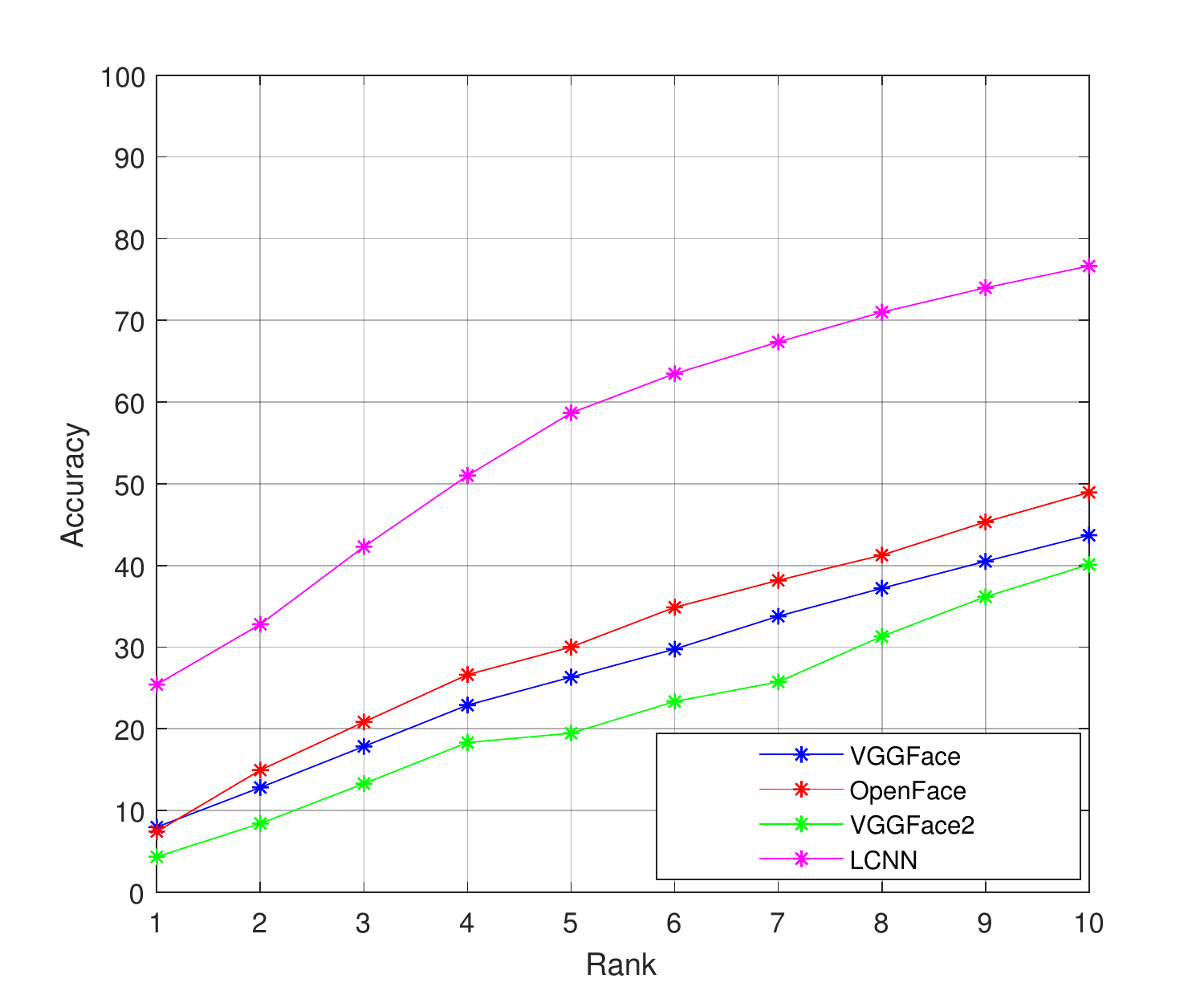}
\caption{CMC curves of the four pre-trained deep face recognition models.}
\label{fig:CMCBaseline}
\end{figure}

\begin{table}[]
\centering
\small
\caption{Comparison of mean inter-class distance of gallery image of each subject with probe images of other subjects using pre-trained LCNN-9, CL, TL, and SCL in the testing set.}
\begin{tabular}{|c|c|c|c|}
\hline
\textbf{LCNN-9} & \textbf{CL} & \textbf{TL} & \textbf{SCL} \\ \hline
0.81           & 0.86       & 0.84      & 0.91        \\ \hline
\end{tabular}
\label{tab:InterClassDist}
\end{table}

\subsection{Injured Face Recognition using Pre-trained Models}
In order to evaluate the performance of pre-trained models on the IF database, four existing pre-trained deep face recognition models, namely, VGG-Face, OpenFace, VGGFace2, and LightCNN-9 are used. Features are extracted using the pre-trained models from the gallery and probe of the testing set. Extracted features are used to match the probe images with the gallery images using Euclidean distance. Table \ref{BaselineAcc} shows the mean classification accuracy with standard deviation of five times random subsampling based cross-validation at rank 1, rank 5, and rank 10. Figure \ref{fig:CMCBaseline} shows the Cumulative Match Characteristic (CMC) curve. It is observed that existing face recognition models do not perform well for the task of injured face image recognition. Among the pre-trained models, LightCNN-9 performs best with an average classification accuracy of 25.43\% at rank 1.  


\subsection{Injured Face Recognition using Subclass Contrastive Loss}
\begin{figure}[t]
\centering
\includegraphics[scale = 0.75]{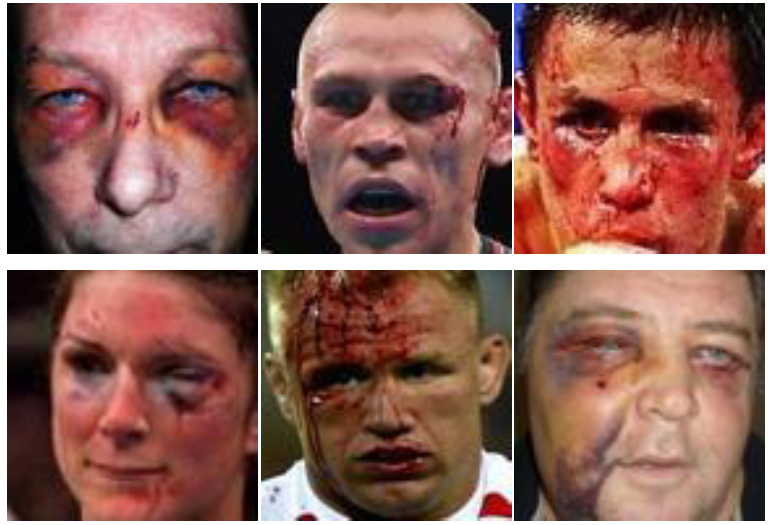}
\caption{Samples misclassified by CL but are correctly classified by SCL.}
\label{fig:Misclassified_Collage}
\end{figure}

This experiment is performed to evaluate the performance of the proposed algorithm with SCL, and the results are compared with CL (Contrastive  Loss) and TL (Triplet Loss). Among the baseline models, LightCNN-9 performs the best. Therefore, it is used as the base network to train the proposed SCL, CL, and TL. Table \ref{ClassAccuracy} shows the mean classification accuracy with standard deviation at rank 1, 5, and 10 obtained from these algorithms. It is observed that SCL outperforms both CL and TL. For instance, SCL yields 8.03\% and 6.44\% higher rank-1 accuracies than CL and TL, respectively. The low performance of CL and TL indicates the limitation of existing networks in handling large intra-class variations among the injured faces of the same subject and small inter-class distance among different subjects. This limitation is overcome by the proposed algorithm with SCL as these variations are explicitly encoded in the proposed loss function. Table \ref{tab:InterClassDist} shows the comparison of the mean inter-class distance of gallery image of each subject with probe images of other subjects using pre-trained LCNN-9, CL, TL, and SCL. It is observed that training with SCL increases the inter-class separation, which in turn results in the improved performance. Figure \ref{fig:Misclassified_Collage} shows some samples misclassified by the CL based approach but they are correctly classified by the proposed SCL based model.





\begin{figure}[t]
\centering
\includegraphics[scale = 0.425]{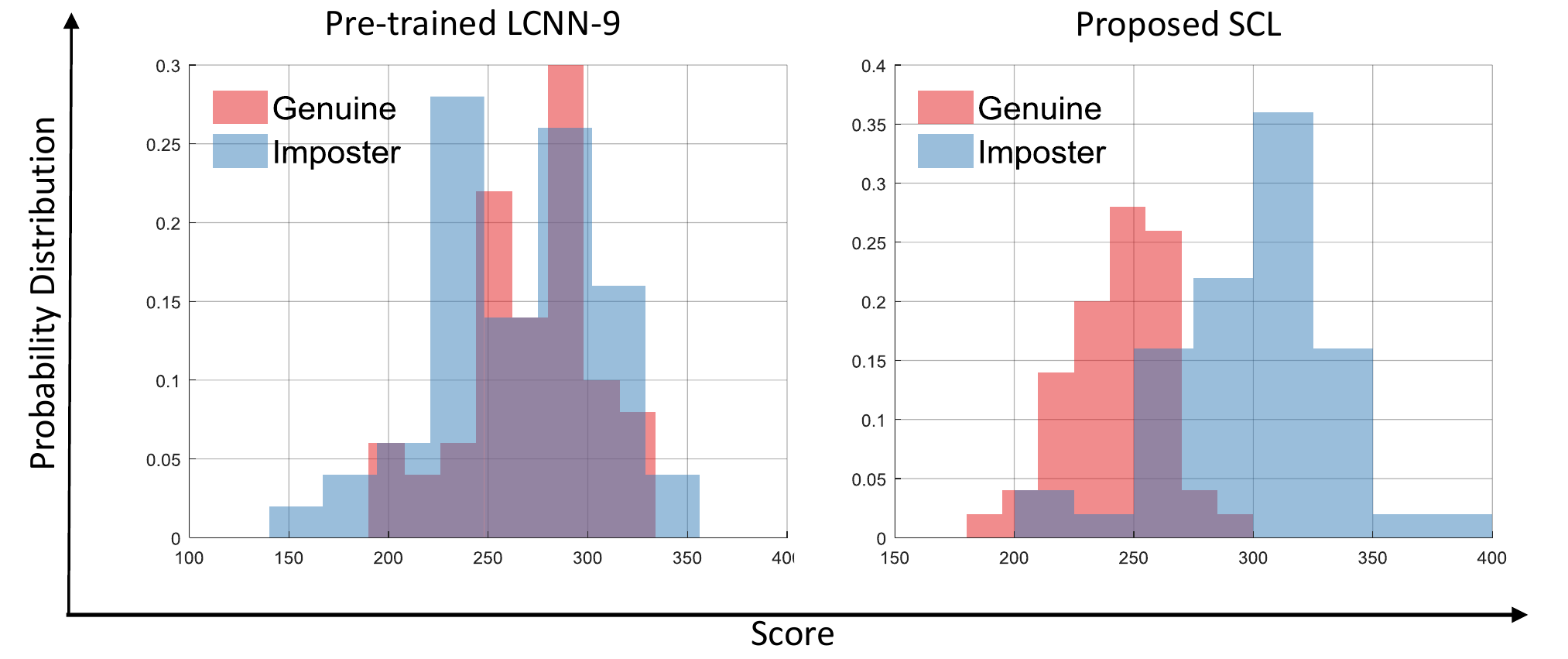}
\caption{Genuine and imposter score distributions using pre-trained LCNN-9 and proposed SCL.}
\label{fig:ScoreDistribution}
\end{figure}

Verification performance is also evaluated on the IF database using CL, TL, and SCL. For this purpose, 50 genuine and 50 imposter pairs are generated. It is observed that the proposed SCL performs better than CL and TL for the task of verification. For instance, at 0.1 False Accept Rate (FAR), the Genuine Accept Rate (GAR) is 74.00\%, 76.00\%, and 78.00\% corresponding to CL, TL, and SCL. Figure \ref{fig:ScoreDistribution} shows the genuine and imposter score distribution using pre-trained LightCNN-9 and SCL, respectively. It is observed that the proposed SCL reduces the overlap between the two distributions, which in turn results in the improved performance.  
 
\subsection{Extended Gallery Experiment}
The proposed algorithm with SCL is also evaluated for extended gallery experiment. In face recognition, extended gallery experiment is performed to emulate the real-world scenario of matching the probe images with a large gallery \cite{nagpal2017face}. Therefore, an additional 100 subjects with a single image per subject from the SCface dataset are added to the gallery of the testing set to evaluate the performance of the proposed SCL. The SCface dataset contains 4160 static images of 130 subjects in the visible and infrared spectrum. For the experimental purpose, images of the visible spectrum are used. The model trained on the training set of the IF database is used for evaluation. Table \ref{ClassAccuracyExtendedGal} shows the comparison of the classification accuracy obtained with and without an extended gallery. It is observed that the proposed algorithm with SCL performs equally well in extended gallery experiment.

\begin{table}[]
\small
\centering
\caption{Results of the extended gallery experiment. `With EG' stands for with extended gallery experiment and `Without EG' stands for without extended gallery experiment.}
\label{ClassAccuracyExtendedGal}
\begin{tabular}{|c|c|c|c|}
\hline
                    & \textbf{Rank 1}  & \textbf{Rank 5}  & \textbf{Rank 10} \\ \hline
\textbf{With EG}    & 36.13 $\pm$ 3.8 & 65.91 $\pm$ 6.0 & 76.62 $\pm$ 4.5 \\ \hline
\textbf{Without EG} & 36.70 $\pm$ 3.6 & 65.91 $\pm$ 6.0 & 82.4 $\pm$ 6.5 \\ \hline
\end{tabular}
\end{table}

\section{Conclusion}
Facial injuries are common in road accidents, violence, and sports. Facial injuries change the physical structure of the face and damage soft tissues. Several cases of unidentified bodies with facial injuries and unconscious victims of road accidents require timely identification. Therefore, in this research, the problem of injured face recognition is studied. For this purpose, a novel Subclass Contrastive Loss (SCL) is proposed for recognition of injured faces. Additionally, first of its kind Injured Face (IF) database is created and baseline performance of existing deep face recognition models is evaluated on the IF database. Multiple experiments showcase the efficacy of the proposed SCL algorithm for injured face recognition. 

\section*{Acknowledgements}
This research is partly supported by the Infosys Center of Artificial Intelligence, IIIT Delhi, India. M. Vatsa is also partially supported by the Department of Science and Technology, Government of India through Swarnajayanti Fellowship. The authors acknowledge Maneet Singh and Rohit Keshari for their constructive and useful feedback.

{\small
\bibliographystyle{ieee}
\bibliography{submission_example}

\begin{thebibliography}{10}\itemsep=-1pt

\bibitem{WinNT1}
{WHO} statistics.
\newblock \url{http://tinyurl.com/y6dz83t4}.

\bibitem{abaza2009retrieving}
A.~Abaza, A.~Ross, and H.~Ammar.
\newblock Retrieving dental radiographs for post-mortem identification.
\newblock In {\em IEEE International Conference on Image Processing}, pages
  2537--2540, 2009.

\bibitem{amos2016openface}
B.~Amos, B.~Ludwiczuk, J.~Harkes, P.~Pillai, K.~Elgazzar, and
  M.~Satyanarayanan.
\newblock Openface: Face recognition with deep neural networks.
\newblock In {\em IEEE Winter Conference on Applications of Computer Vision},
  2016.

\bibitem{black2017prevention}
A.~M. Black, D.~A. Patton, P.~H. Eliason, and C.~A. Emery.
\newblock Prevention of sport-related facial injuries.
\newblock {\em Clinics in Sports Medicine}, 36(2):257--278, 2017.

\bibitem{bodkha2012role}
P.~Bodkha and B.~Yadav.
\newblock A role of digital imaging in identification of unidentified bodies.
\newblock {\em Journal of Indian Academy of Forensic Medicine},
  34(4):0971--0973, 2012.

\bibitem{canzi2019cfi}
G.~Canzi, E.~De~Ponti, G.~Novelli, F.~Mazzoleni, O.~Chiara, A.~Bozzetti, and
  D.~Sozzi.
\newblock The cfi score: Validation of a new comprehensive severity scoring
  system for facial injuries.
\newblock {\em Journal of Cranio-Maxillofacial Surgery}, 47(3):377--382, 2019.

\bibitem{cao2018vggface2}
Q.~Cao, L.~Shen, W.~Xie, O.~M. Parkhi, and A.~Zisserman.
\newblock Vggface2: A dataset for recognising faces across pose and age.
\newblock In {\em IEEE International Conference on Automatic Face \& Gesture
  Recognition}, pages 67--74, 2018.

\bibitem{clayton1995identification}
T.~Clayton, J.~Whitaker, and C.~Maguire.
\newblock Identification of bodies from the scene of a mass disaster using dna
  amplification of short tandem repeat (str) loci.
\newblock {\em Forensic Science International}, 76(1):7--15, 1995.

\bibitem{coulibaly2018inter}
T.~A. Coulibaly, R.~B{\'e}ogo, I.~Traor{\'e}, H.~M. Kohoun, and B.~V. Ili.
\newblock Inter personal violence-related facial injuries: a 10-year survey.
\newblock {\em Journal of Oral Medicine and Oral Surgery}, 24(1):2--5, 2018.

\bibitem{grgic2011scface}
M.~Grgic, K.~Delac, and S.~Grgic.
\newblock Scface--surveillance cameras face database.
\newblock {\em Multimedia tools and applications}, 51(3):863--879, 2011.

\bibitem{hadsell2006dimensionality}
R.~Hadsell, S.~Chopra, and Y.~LeCun.
\newblock Dimensionality reduction by learning an invariant mapping.
\newblock In {\em IEEE Computer Society Conference on Computer Vision and
  Pattern Recognition}, volume~2, pages 1735--1742, 2006.

\bibitem{jordan2006management}
J.~Jordan and K.~Calhoun.
\newblock Management of soft tissue trauma and auricular trauma.
\newblock {\em Bailey BJ, Johnson JT, Newlands SD. Head \& Neck Surgery:
  Otolaryngology. Hagerstwon, MD: Lippincott Williams \& Wilkins}, pages
  935--36, 2006.

\bibitem{majumdar2018detecting}
P.~Majumdar, S.~Chhabra, R.~Singh, and M.~Vatsa.
\newblock On detecting domestic abuse via faces.
\newblock In {\em IEEE Conference on Computer Vision and Pattern Recognition
  Workshops}, pages 2173--2179, 2018.

\bibitem{nagpal2017face}
S.~Nagpal, M.~Singh, R.~Singh, M.~Vatsa, A.~Noore, and A.~Majumdar.
\newblock Face sketch matching via coupled deep transform learning.
\newblock In {\em Proceedings of the IEEE International Conference on Computer
  Vision}, pages 5419--5428, 2017.

\bibitem{nassar2008automatic}
D.~E. Nassar, A.~Abaza, X.~Li, and H.~Ammar.
\newblock Automatic construction of dental charts for postmortem
  identification.
\newblock {\em IEEE Transactions on Information Forensics and Security},
  3(2):234--246, 2008.

\bibitem{parkhi2015deep}
O.~M. Parkhi, A.~Vedaldi, A.~Zisserman, et~al.
\newblock Deep face recognition.
\newblock In {\em British Machine Vision Conference}, volume~1, page~6, 2015.

\bibitem{quinet2016unclaimed}
K.~Quinet, S.~Nunn, and A.~Ballew.
\newblock Who are the unclaimed dead?
\newblock {\em Journal of Forensic Sciences}, 61:S131--S139, 2016.

\bibitem{schroff2015facenet}
F.~Schroff, D.~Kalenichenko, and J.~Philbin.
\newblock Facenet: A unified embedding for face recognition and clustering.
\newblock In {\em IEEE Conference on Computer Vision and Pattern Recognition},
  pages 815--823, 2015.

\bibitem{singh2012maxillofacial}
V.~Singh, L.~Malkunje, S.~Mohammad, N.~Singh, S.~Dhasmana, and S.~K. Das.
\newblock The maxillofacial injuries: A study.
\newblock {\em National Journal of Maxillofacial Surgery}, 3(2):166, 2012.

\bibitem{trokielewicz2016post}
M.~Trokielewicz, A.~Czajka, and P.~Maciejewicz.
\newblock Post-mortem human iris recognition.
\newblock In {\em IEEE International Conference on Biometrics}, pages 1--6,
  2016.

\bibitem{trokielewicz2019iris}
M.~Trokielewicz, A.~Czajka, and P.~Maciejewicz.
\newblock Iris recognition after death.
\newblock {\em IEEE Transactions on Information Forensics and Security},
  14(6):1501--1514, 2019.

\bibitem{wu2018light}
X.~Wu, R.~He, Z.~Sun, and T.~Tan.
\newblock A light cnn for deep face representation with noisy labels.
\newblock {\em IEEE Transactions on Information Forensics and Security},
  13(11):2884--2896, 2018.

\end{thebibliography}
}

\end{document}